\documentclass[letterpaper]{article} 
\usepackage{aaai2026}  
\usepackage{times}  
\usepackage{helvet}  
\usepackage{courier}  
\usepackage[hyphens]{url}  
\usepackage{graphicx} 
\usepackage{amsmath} 
\usepackage{booktabs} 
\usepackage{makecell} 
\usepackage{amssymb} 

\usepackage{natbib}  
\usepackage{caption} 
\usepackage{todonotes}

\usepackage{algorithm}
\usepackage{algorithmic}

\usepackage{newfloat}
\usepackage{listings}
\DeclareCaptionStyle{ruled}{labelfont=normalfont,labelsep=colon,strut=off} 
\floatstyle{ruled}
\newfloat{listing}{tb}{lst}{}
\floatname{listing}{Listing}
\nocopyright

\title{Methodological Framework for Quantifying Semantic Test Coverage in RAG Systems}
\author{
    Noah Broestl\textsuperscript{\rm 1}
    Max Struever\textsuperscript{\rm 1}
    Adel Nasser Abdalla\textsuperscript{\rm 1}
    Hersh Gupta\textsuperscript{\rm 1}
    Rajprakash Bale\textsuperscript{\rm 1}
}

\affiliations{
    \textsuperscript{\rm 1}Boston Consulting Group
}

\begin{document}

\maketitle
\begin{abstract}
Reliably determining the performance of Retrieval-Augmented Generation (RAG) systems depends on comprehensive test questions. While a proliferation of evaluation frameworks for LLM-powered applications exists, current practices lack a systematic method to ensure these test sets adequately cover the underlying knowledge base, leaving developers with significant blind spots. To address this, we present a novel, applied methodology to quantify the semantic coverage of RAG test questions against their underlying documents. Our approach leverages existing technologies, including vector embeddings and clustering algorithms, to create a practical framework for validating test comprehensiveness. Our methodology embeds document chunks and test questions into a unified vector space, enabling the calculation of multiple coverage metrics: basic proximity, content-weighted coverage, and multi-topic question coverage. Furthermore, we incorporate outlier detection to filter irrelevant questions, allowing for the refinement of test sets. Experimental evidence from two distinct use cases demonstrates that our framework effectively quantifies test coverage, identifies specific content areas with inadequate representation, and provides concrete recommendations for generating new, high-value test questions. This work provides RAG developers with essential tools to build more robust test suites, thereby improving system reliability and extending to applications such as identifying misaligned documents.
\end{abstract}

\section{Introduction}
Large Language Models (LLMs) have become a foundational technology across multiple domains, with Retrieval-Augmented Generation (RAG) systems emerging as a practical and widely adopted LLM application pattern \cite{rwrag,rag-survey}. RAG systems enhance LLMs by retrieving relevant information from a knowledge base before generating responses, thereby addressing limitations in factual accuracy and domain-specific knowledge.

RAG system adoption has rapidly accelerated across industries due to several key advantages. First, RAG allows organizations to leverage private or domain-specific knowledge bases without costly model fine-tuning. Even when fine-tuning is pursued, pure RAG systems often outperform fine-tuned alternatives \cite{ragvsfinetune,finetuningretrievalcomparing}. Second, RAG systems offer greater transparency and traceability by citing source documents used for response generation \cite{rag-survey}. Third, RAG helps mitigate hallucination by grounding model responses in retrieved factual context \cite{ingestandground,hallucinationsurvey}.

As these systems become increasingly critical to business operations, the need for reliable performance intensifies. Incorrect information from a medical RAG system could jeopardize patient care, while a legal RAG system might expose an organization to compliance risks. These high-stakes applications underscore the importance of comprehensive testing.

Despite significant progress in tools and frameworks for evaluating RAG system outputs and retrieval quality, a critical gap persists: assessing whether test questions adequately cover the underlying knowledge base. This blind spot raises a fundamental question: How can we ensure our test set sufficiently covers the information space we aim to evaluate? 

There are many methods to evaluate the components of a RAG system. Each method requires a test set, which can be generated by human experts, humans aided by automation, or entirely by automation \cite{hcat}. Regardless of the generation method, understanding the test set's coverage of system behaviors is crucial. Especially with LLMs, where general capabilities and non-determinism make exhaustive testing impossible, understanding what is being tested and identifying areas lacking insight is a foundational concern.

While human expertise is essential for determining test coverage, fully automated methods for test set generation can create blind spots. For instance, if irrelevant documents are mistakenly uploaded, automatically generated tests could consume resources testing non-existent system components and obfuscate the system performance in the desired domain. However, humans alone cannot manage the scale required for comprehensive RAG testing. Expert judgment, though important, is inherently subjective, limited in scalability, and insufficient to guarantee comprehensive coverage on its own \cite{llmstrustedevaluatingrag,ragdiligence}. Without a systematic evaluation method, organizations cannot confidently answer fundamental questions such as: "Are we truly testing the most important information in our knowledge base?" or "Are there major blind spots in our test suite?"

In traditional software development, code coverage metrics help developers understand the thoroughness of their test suites. A well-understood metric like line coverage shows the percentage of code lines executed during tests, providing guidance for improving test quality and confidence in system reliability \cite{codecoverage, Ellims2004UnitTI}. For example, discovering that a critical payment processing module has only 40\% coverage immediately highlights a risk area and directs testing efforts. Without coverage metrics, teams might have a false sense of security about their test suite's thoroughness.

Teams developing test sets for RAG applications face the following questions:
\begin{itemize}
    \item Do our test questions cover all key topics in our knowledge base?
    \item Are some important conceptual areas underrepresented in our testing?
    \item If users ask about topics in certain areas, how confident can we be that the system will perform correctly?
\end{itemize}

Without a systematic method to assess test coverage, organizations risk unknowingly deploying RAG systems with significant evaluation blind spots, potentially missing critical failure modes or performance issues in production. For instance, a financial advice RAG system might be thoroughly tested on tax-related questions but have minimal testing on retirement planning, creating an unidentified risk area. Our work addresses this gap by developing a methodology to quantify RAG system test coverage, providing concrete metrics and actionable insights to improve test quality.

\section{Related Work}
While test coverage for Retrieval-Augmented Generation (RAG) systems remains relatively unexplored, several related areas have informed our approach to coverage testing for LLMs. 

\subsection{General LLM Evaluation Frameworks}
The field of LLM evaluation has seen rapid development in frameworks and metrics. Holistic Evaluation of Language Models (HELM) \cite{helm} offers a comprehensive benchmarking framework, employing a multi-metric approach to assess accuracy, calibration, robustness, fairness, bias, toxicity, and efficiency across 16 core scenarios. This provides a robust framework for overall model evaluation.

Similarly, BIG-bench \cite{bigbench} presents over 204 diverse tasks, from elementary mathematics to complex reasoning, probing the boundaries of LLM capabilities. The diversity of BIG-bench underscores the need for broad evaluation coverage, though its primary focus remains on general capabilities rather than domain-specific knowledge retrieval.

Other notable benchmarks include Massive Multitask Language Understanding (MMLU) \cite{mmlu}, which evaluates models across 57 subjects (from STEM to humanities), and TruthfulQA \cite{truthfulqa}, which specifically targets model truthfulness. While these frameworks have significantly advanced our understanding of LLM capabilities, they focus on evaluating model \textit{outputs} rather than the \textit{quality of test inputs}.

\subsection{RAG-Specific Evaluation}
For retrieval--augmented generation (RAG) evaluation, two recent frameworks are particularly influential: 
\textbf{RAGAS}~\cite{ragas} and \textbf{ARES}~\cite{ares}.  
RAGAS introduces an end-to-end pipeline for automatically assessing RAG systems through language-model–based judges that approximate human ratings on several axes, including \emph{answer relevance}, \emph{faithfulness to the retrieved context}, and \emph{context precision/recall}.  
ARES tackles a complementary stage of the stack: it fine-tunes compact language-model judges to score responses for context relevance and correctness, and—crucially—can \emph{generate synthetic queries and answers} from the underlying documents.  In practice, this makes ARES useful for stress-testing retrieval pipelines before any production traffic is available, as it can quickly bootstrap large evaluation sets that mimic a closed-book QA setting.

Despite their strengths, both frameworks share two critical limitations.
\textbf{First}, they generate evaluation queries directly from the documents instead of being based on realistic user interactions, so the resulting test sets measure what the corpus can say rather than what users might actually ask; this risks rewarding systems that excel at document-centric questions while overlooking gaps that matter to end-users. 
\textbf{Second}, neither RAGAS nor ARES flags or penalizes irrelevant, duplicate, or out-of-scope documents in the index, allowing a system to score highly even when its data store contains substantial noise. Our approach does not focus on generating new questions; instead, it evaluates users' existing test-set questions and audits the index to surface and mitigate irrelevant documents and questions. Allowing for a user-driven evaluation suite.

While advances in LLM and RAG evaluation are important, a significant gap remains: there is no established methodology for assessing how comprehensively a set of test questions covers the knowledge domain of a RAG system. This gap is particularly notable given the increasing interest in building and deploying RAG systems across domains, as evidenced by surveys showing a growing number of methods \cite{rag-survey} and domain-specific applications \cite{rag-domain-specific}. Our work fills this gap by introducing a systematic approach to measuring test coverage in RAG systems, complementing existing evaluation frameworks and providing a missing piece in the RAG evaluation pipeline.

\section{Methodology}
We propose a novel methodology for measuring test coverage in Retrieval-Augmented Generation (RAG) systems using vector embeddings. This approach quantifies how thoroughly a test set covers the underlying document space by leveraging semantic similarity between document chunks, clusters, and test questions. It also analyzes the relevance of individual test questions, ensuring the test set accurately reflects the document content.

It is crucial to note that the goal of this methodology is to provide a directional assessment of test coverage, rather than evaluating towards some universally defined "ideal" coverage. Ideal coverage is inherently subjective and highly dependent on the specific use case and domain. Instead, our method aims to reveal comparative coverage gaps among conceptual areas within the knowledge base and provide a quantifiable score that allows for the relative comparison of two different test sets. This enables practitioners to understand where their current testing efforts might be lacking relative to the document corpus, and to iteratively improve their test suites.

\subsection{Conceptual Framework}
Our approach projects both document chunks and test questions into the same high-dimensional embedding space. By analyzing their relative positions, we can measure how effectively the test questions span the semantic space of the document corpus.

The core intuition is that well-designed test questions should be semantically "close" to relevant clusters of document chunks in this embedding space. This proximity ensures that key semantic areas within the document corpus are addressed by at least one test question. Conversely, regions in the embedding space containing document chunks but no nearby test questions indicate potential blind spots in the test coverage. 

\begin{figure}
    \centering
    \includegraphics[width=1\linewidth]{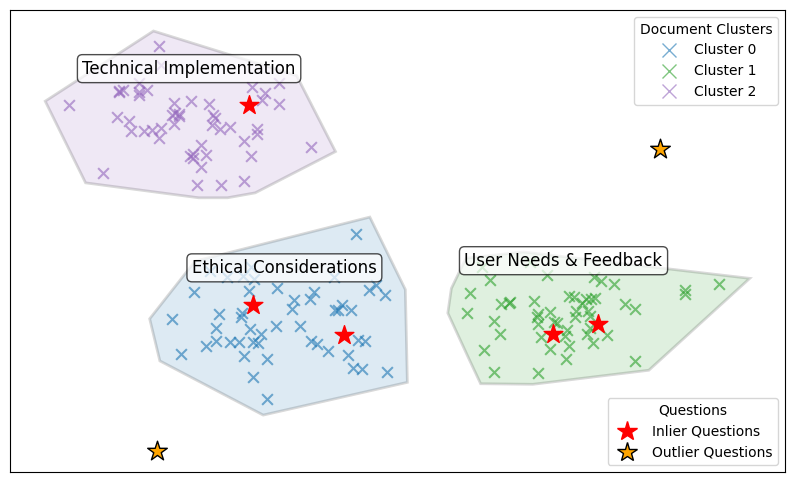}
    \caption{\textbf{Relating Semantic Regions to Test Questions for Coverage and Outlier Identification.} Document chunks ('x') form distinct semantic regions (clusters), with test coverage increasing as each region is addressed by at least one test question. Outliers are questions that are identified as outside of the relevant semantic regions.}
    \label{fig:semantic-visual}
\end{figure}

From a geometric perspective, each embedded document chunk can be viewed as a point in a high-dimensional space, with clusters of chunks forming "regions" of related content. Ideally, test questions should be distributed across these regions to ensure comprehensive coverage. Questions that are excessively clustered or positioned far from relevant document chunks suggest inefficiencies or gaps in the test suite. Figure~\ref{fig:semantic-visual} illustrates these concepts, highlighting clusters with coverage and outlier questions.

Before calculating coverage, it's crucial to identify and remove irrelevant questions from the test set. These are questions semantically unrelated to the document content, and their inclusion would distort coverage metrics. We achieve this using the Local Outlier Factor (LOF) algorithm \cite{lof}, which identifies such outliers based on their semantic distance from the main document embeddings. This ensures our evaluation focuses only on questions truly representative of the document content.

\subsection{Mathematical Formulation}
\begin{itemize}
    \item Let \( D = \{d_1, d_2, \ldots, d_n\} \) be the set of document chunks.
    
    \item Let \( Q = \{q_1, q_2, \ldots, q_m\} \) be the set of test questions.
    
    \item Let \( Q^\alpha \subseteq Q \) denote the subset of questions retained according to their LOF (Local Outlier Factor) scores. These are the questions considered *in-distribution* or representative, as determined by thresholding the LOF scores.
    
    \item Let \( E_D \in \mathbb{R}^{n \times k} \) be the matrix of document chunk embeddings, where each row \( E_D[i] \) is the \( k \)-dimensional embedding of chunk \( d_i \).
    
    \item Let \( E_Q \in \mathbb{R}^{|Q^\alpha| \times k} \) be the matrix of question embeddings corresponding to \( Q^\alpha \), where each row \( E_Q[j] \) is the embedding of question \( q_j \in Q^\alpha \).
    
    \item Let \( \text{dist}(u, v) \) be a distance function between vectors (we use cosine distance), defined as:
    \[
    \text{dist}(u, v) = 1 - \frac{u \cdot v}{\|u\| \cdot \|v\|}
    \]
\end{itemize}

\subsection{Identifying Outlier Questions}
Before computing coverage metrics, it is essential to identify and filter out irrelevant questions—those semantically dissimilar to the document content. Their inclusion would skew coverage metrics, leading to inaccurate results.

We employed the Local Outlier Factor (LOF) algorithm \cite{lof}, an unsupervised anomaly detection method that assesses the local density of data points relative to their neighbors. In our context, each question is embedded into a high-dimensional vector space, and LOF is applied to these embeddings to identify questions residing in low-density regions. These semantically isolated questions are deemed irrelevant to the document subject matter and excluded from subsequent coverage calculations. This ensures our evaluation focuses solely on questions representative of the document content. Users are provided with an LOF score for each question, indicating the likelihood of it being an outlier. Higher positive scores suggest a higher probability of being an outlier, while negative or near-zero scores indicate an inlier.

As shown in Figure~\ref{fig:semantic-visual}, questions semantically dissimilar to the document corpus are flagged as outliers by the LOF algorithm. Moving forward, all coverage metrics utilize only these "inlier" questions.

\subsection{Coverage Metrics}
We define multiple coverage metrics to handle a variety of use cases.

\subsubsection{Basic Coverage}

The basic coverage metric measures the average proximity of document chunks to the nearest test question. Formally, for each document chunk \( d_i \) with embedding \( E_D[i] \), and each 'inlier' test question \( Q^\alpha = \{q_1, q_2, \ldots, q_{m'}\} \subseteq Q^\alpha \) we find the minimum distance as:

\begin{align*}
\text{mindist}(d_i) &= \min_{j \in \{1, 2, \ldots, m'\}} \text{dist}(E_D[i], E_Q[j]), \quad q_j \in Q^\alpha \\
\intertext{The basic coverage is then:}
eC_{\text{basic}} &= 1 - \frac{1}{n} \sum_{i=1}^{n} \text{mindist}(d_i)
\end{align*}

For interpretability (e.g. higher scores are better), the formula converts distances to similarities by subtracting from 1. It is important to note that this score is not absolute. A score of 100\% would indicate complete syntactic alignment between the test set and the document chunks, an undesirable state. Instead it is a relative measure, allowing one to compare one set of test questions to another and thus needs to be interpreted beyond the raw score.

\subsubsection{Weighted Coverage}

The weighted coverage metric places greater emphasis on document clusters with more content, ensuring that semantically rich areas of the corpus are more thoroughly tested.

We cluster the document chunks into \( K \) clusters using K-means clustering \(C = \{ C_1, C_2, \dots, C_K \}\) where each \( C_k \) is a set of indices corresponding to document chunks in that cluster. For each cluster \( C_k \), we calculate the average minimum cluster distance as:

\begin{align*}
\text{clusterdist}(C_k) = \frac{1}{|C_k|} \sum_{i \in C_k} \text{mindist}(d_i) \\
\intertext{The weighted coverage is then:}
eC_{\text{weighted}} = \sum_{k=1}^{K} \frac{|C_k|}{n} \cdot \left(1 - \text{clusterdist}(C_k)\right)
\end{align*}

This weights each cluster's coverage by its relative size, ensuring that larger semantic regions contribute more to the overall score.

\subsubsection{Multi-Cluster Coverage}

To account for the fact that certain questions may target multiple semantic cluster, the multi-cluster coverage metric allows for questions to be associated with multiple clusters as defined by a configurable distance threshold.

For each test question \( q^\alpha \) and cluster centroid \( c_k \), the question covers the cluster if:

\[
\text{dist}(E_Q[\alpha], c_k) < \text{threshold}
\]

We define the set of clusters covered by each question as:

\[
\text{Cov}_{\text{thresh}}(q^\alpha) = \left\{ C_k \mid \text{dist}(E_Q[\alpha], c_k) < \text{threshold} \right\}
\]

 Then, for each cluster \( C_k \), we define the set of covering questions as:

\[
Q_k = \left\{ q^\alpha \mid C_k \in \text{Cov}_{\text{thresh}}(q^\alpha) \right\}
\]

If \( Q_k \neq \emptyset \), we compute:

\[
\text{multidist}(C_k, Q_k) = \frac{1}{|C_k|} \sum_{i \in C_k} \min_{q^\alpha \in Q_k} \text{dist}(E_D[i], E_Q[\alpha])
\]

Finally, the threshold-based multi-coverage score is:

\[
C_{\text{multi-thresh}} = \sum_{k=1}^{K} \frac{|C_k|}{n} \cdot \left(1 - \text{multidist}(C_k, Q_k)\right)
\]

Clusters with no covering questions contribute 0 to the score. Meaningful thresholds have been observed between .15 and .8 depending on technical details and use case.

\section{Implementation}
Our methodology integrates the proposed coverage metrics into a comprehensive, systematic Python workflow for evaluating RAG test sets, detailed as follows:

\begin{enumerate}
    \item \textbf{Document Chunking:} The document corpus is split into semantically meaningful chunks using a recursive character text splitter with configurable chunk size and overlap.
    
    \item \textbf{Embedding:} Both document chunks and test questions are transformed into high-dimensional vector embeddings using state-of-the-art models (e.g., OpenAI, Voyage AI). 
    
    \item \textbf{Clustering:} Document chunks are clustered using K-means to identify major semantic areas. The number of clusters is a configurable parameter, with a default value determined heuristically based on the corpus size.

    \item \textbf{Outlier Identification:} We leverage the LOF algorithm to detect which questions in the test set are relevant to the documents. Any question deemed irrelevant are filtered out and not considered in downstream calculations. Additionally, these questions are flagged such that the user for review and action.
    
    \item \textbf{Distance Calculation:} We calculate the pairwise distances between filtered test questions and document chunks/clusters using pairwise-cosine distance, which is well-suited for comparing semantic similarities in embedding space.
    
    \item \textbf{Coverage Calculation:} The defined metrics are calculated based on these distances, providing multiple perspectives on test coverage.
    
    \item \textbf{Gap Analysis:} Areas of low coverage are identified by finding clusters with high minimum distances to any test question. An LLM is then used to analyze the clusters, and extract key themes along with suggested questions, such that the user may integrate additional questions to close the coverage gap in these thematic areas.
    
\end{enumerate}

This is defined formally in Algorithm \ref{alg:coverage}. Components supporting the core coverage metric are embeddings, gap analysis, and visualization.

\begin{algorithm}
\caption{RAG Test Coverage Calculation}
\begin{algorithmic}
    \REQUIRE Document corpus \( D \)
    \REQUIRE Test questions \( Q \)
    \REQUIRE Parameters (chunk\_size, num\_clusters, etc.)
    \ENSURE Coverage metrics, efficiency scores, and identified coverage gaps

    \textbf{1: Data Preparation and Embedding Generation}
    \STATE Chunk the document corpus \( D \) into \( n \) chunks \( \{d_1, d_2, \ldots, d_n\} \)
    \STATE Generate embeddings \( E_D \) for document chunks and \( E_Q \) for test questions

    \textbf{2: Semantic Space Organization}
    \STATE Cluster document embeddings \( E_D \) into \( K \) clusters using K-means

    \textbf{3: Question Filtering and Relevance Assessment}
    \STATE Apply Local Outlier Factor (LOF) to \( E_Q \) to identify in-distribution questions:
    \begin{ALC@g}
        \STATE a. Calculate LOF score for each question embedding \( E_Q[j] \)
        \STATE b. Filter \( Q \) to create \( Q^\alpha \subseteq Q \), containing only in-distribution (non-outlier) questions, which will be used for downstream calculations.
    \end{ALC@g}

    \textbf{4: Core Coverage Metric Calculations using \(Q^\alpha\)}
    \STATE Calculate basic coverage \( C_{\text{basic}} \):
    \begin{ALC@g}
        \STATE a. For each document chunk \( d_i \), find $\min_{q_j \in Q^\alpha} \text{dist}(E_D[i], E_Q[j])$
        \STATE b. Average these minimum distances and convert to a similarity score ($1 - \text{average\_mindist}$)
    \end{ALC@g}

    \STATE Calculate weighted coverage \( C_{\text{weighted}} \):
    \begin{ALC@g}
        \STATE a. For each cluster \( C_k \), calculate average minimum distance of its chunks to \( Q^\alpha \)
        \STATE b. Weight by cluster size \( |C_k|/n \) and sum across clusters
    \end{ALC@g}

    \STATE Calculate multi-coverage \( C_{\text{multi-N}} \):
    \begin{ALC@g}
        \STATE a. For each \( q^\alpha \in Q^\alpha \), identify $\text{Cov}_N(q^\alpha)$ (N closest clusters)
        \STATE b. For each cluster \( C_k \), calculate average minimum distance of its chunks to covering questions in \( Q_k \)
        \STATE c. Aggregate weighted by cluster size; clusters with no covering questions contribute 0
    \end{ALC@g}

    \textbf{5: Gap Analysis and Recommendation}
    \STATE Identify coverage gaps:
    \begin{ALC@g}
        \STATE a. Find clusters with coverage below a defined threshold
        \STATE b. Extract key themes and suggested questions from these clusters using an LLM
        \STATE c. Sort by coverage score and size to prioritize
    \end{ALC@g}

    \STATE \textbf{return} coverage metrics, efficiency scores, and identified gaps

\end{algorithmic}
\label{alg:coverage}
\end{algorithm}

\subsection{Embeddings}

Our implementation supports multiple embedding providers to accommodate different user preferences and requirements. We use OpenAI and Voyage as providers, including OpenAI’s \texttt{text-embedding-3-small}, \allowbreak \texttt{embedding-3-large}, and \allowbreak \texttt{ada-002}, as well as Voyage’s \texttt{voyage-3}.

The embedding models were selected based on their proven performance in semantic similarity tasks. The OpenAI \texttt{embedding-3-large} model, for instance, achieves state-of-the-art performance on the MTEB (Massive Text Embedding Benchmark) with a score of 65.4\%, outperforming other publicly available models. The Voyage AI \texttt{voyage-3} model offers a balance between quality and computational efficiency, with strong performance on retrieval tasks.

It is important to standardize the embedding model that is used. While \cite{vec2vec} shows that large enough embeddings will converge to the same latent representation, the proximity of said embeddings can differ. In other words certain embedding models display tighter semantic clustering, grouping related words more closely together in latent space, which can lead to more pronounced local neighborhood structures.

\subsection{Gap Analysis}

Beyond providing coverage metrics, our implementation provides actionable insights by identifying specific areas of the document corpus with low coverage and characterizes them using thematic analysis. This enables users to target test creation efforts toward the most significant gaps.

The gap analysis process involves:

\begin{enumerate}
    \item \textbf{Identifying low-coverage clusters:} Clusters with coverage scores below a configurable threshold (default: 0.7) are identified as gaps.
    
    \item \textbf{Extracting key themes:} For each low-coverage cluster, we extract representative themes using LLM-based concept extraction. This leverages GPT models to analyze the content and extract 3--5 key concepts per cluster.
    
    \item \textbf{Ranking gaps by importance:} Gaps are ranked based on a combination of their coverage score and the size of the cluster, prioritizing large clusters with poor coverage.
    
    \item \textbf{Generating recommendations:} The system provides actionable recommendations for improving test coverage, including suggested themes and new test questions.
\end{enumerate}

\noindent The output of this analysis for a low coverage cluster is shown in Table~\ref{tab:gap-analysis}.

\begin{table}[h!]
\centering
\caption{Gap Analysis Summary for Cluster 2}
\label{tab:gap-analysis}
\begin{tabular}{@{}ll@{}}
\toprule
\textbf{Cluster ID}         & 2 \\
\textbf{Coverage Score}     & 0.42 \\
\textbf{Cluster Size}       & 15 documents \\
\textbf{Corpus Share}       & 12.5\% \\
\textbf{Extracted Themes}   & \makecell[l]{-- Data Privacy\\-- User Consent\\-- Information Storage} \\
\textbf{Suggested Question} & What are some privacy best practices? \\
\bottomrule
\end{tabular}
\end{table}

\noindent Table~\ref{tab:gap-analysis} indicates that cluster 2, which represents 12.5\% of the document corpus and focuses on themes of Data Privacy, User Consent, and Information Storage, has a low coverage score of 0.42 and should be prioritized for additional test questions (potentially adding the ones that are suggested).

\subsection{Visualization}

To aid in interpreting coverage results, our implementation includes visualization capabilities using t-SNE (t-Distributed Stochastic Neighbor Embedding) \cite{tsne} for dimensionality reduction. This allows users to visualize the high-dimensional embedding space in a 2D plot, showing the relationship between document chunks and test questions.

Document chunks are colored by cluster and test questions are highlighted as prominent markers. The plots include annotations for coverage metrics and can be used to intuitively identify areas of poor coverage.

\section{Real World Validation}
\label{sec:experiments}

To validate the practical utility and robustness of our proposed coverage framework, we apply it to two distinct real-world scenarios. Each scenario highlights a common challenge in RAG system deployment and demonstrates how our methodology effectively identifies and addresses these limitations, thereby improving the reliability of the system.

\subsection{Use Case Descriptions}

\begin{figure}
    \centering
    \includegraphics[width=1\linewidth]{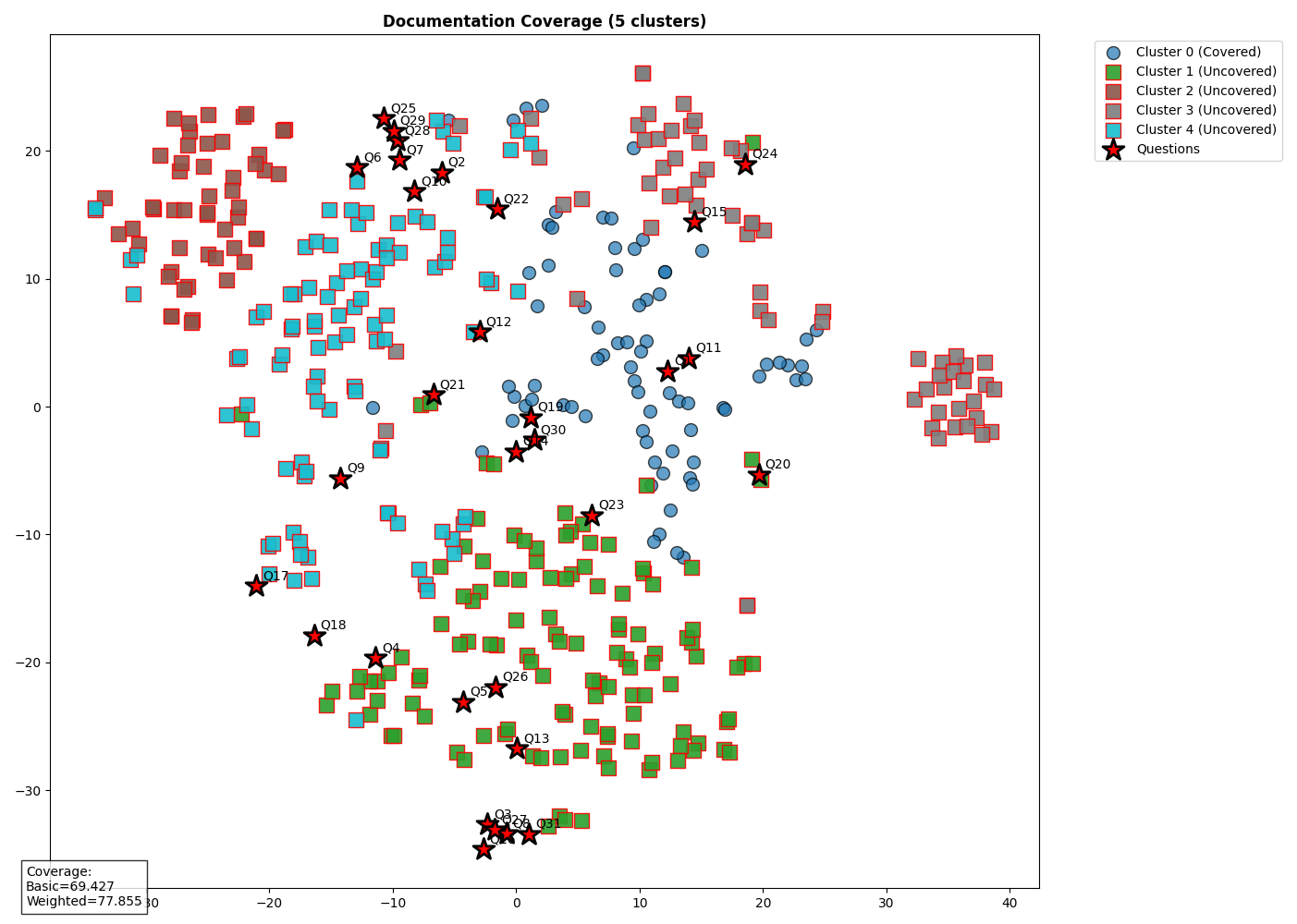}
    \caption{\textbf{Visualization of initial test coverage for a real-world product use case.}}
    \label{fig:coverage_before}
\end{figure}

The first use case is a RAG system deployed in a real-world product with comprehensive documentation about system usage. In our initial evaluation of test set quality, the documentation dataset, comprising 415 chunks (each 500 tokens), along with an existing set of 31 test questions provided by the product team, served as a representative artifact for evaluation. Initial application of our methodology to this corpus revealed significant limitations in the existing test coverage. The overall basic coverage was a low 69.4\%, with four out of five semantic document clusters demonstrating blind spots in conceptual clusters (e.g., 'Talent conflicts \& Slating status,' 'Org-wide structure metrics \& dashboards') as illustrated in Figure~\ref{fig:coverage_before}. This real-world scenario clearly presented a need for more comprehensive evaluation of its RAG system's test suite.

The second scenario focused on evaluating the impact of irrelevant content. We constructed a corpus using a Schwab S\&P 500 Index Fund prospectus as the primary relevant document, alongside five Schwab-specific questions. To simulate the challenge of managing extraneous information in a real system, we deliberately injected a separate, semantically unrelated document describing various bird species into this corpus. The combined documents were chunked into 55 segments (500 tokens each) and clustered into three distinct topical groups.

In both scenarios, document chunks and test questions were processed by generating embeddings with the same model, and document chunks were subsequently clustered using K-means, forming the foundational input for our coverage analysis.

\subsection{Addressing Challenge 1: Incomplete Test Coverage}
\begin{figure}
    \centering
    \includegraphics[width=1\linewidth]{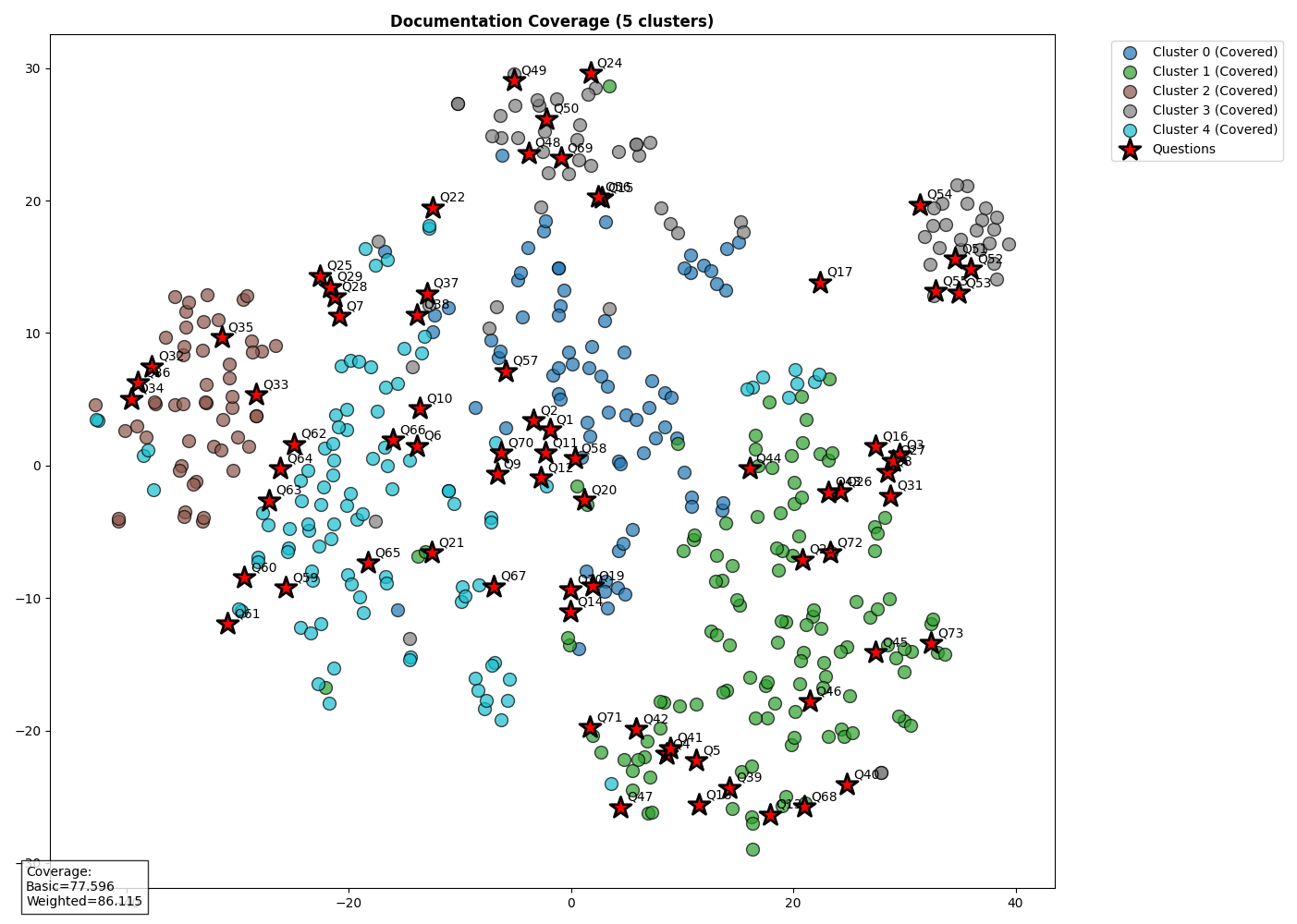}
    \caption{\textbf{Visualization of test coverage for a real-world product use case after adding questions to the indicated clusters.}}
    \label{fig:coverage_after}
\end{figure}

The first use case directly illustrates how our methodology addresses the common real-world challenge of incomplete test coverage. The initial analysis clearly signaled that the existing 31 questions lacked sufficient representation across key semantic areas of the documentation. By leveraging the themes of the uncovered clusters identified by our framework, additional test questions were generated using an LLM. The inclusion of these new, thematically aligned questions significantly improved the basic coverage to 77.6\% and resulted in all document clusters showing some association with tests, as illustrated in Figure~\ref{fig:coverage_after}. This demonstrates our framework's practical utility as a diagnostic tool for detecting sparse or insufficiently targeted question sets within real-world RAG systems, guiding administrators to expand and improve their evaluation questions effectively. The observed positive shift in coverage reflects a more comprehensive and better-aligned question set.

\subsection{Addressing Challenge 2: Presence of Irrelevant Knowledge}
Our second use case addresses another critical real-world challenge: the presence of semantically misaligned or irrelevant documents within a RAG system's knowledge base. To demonstrate this, we constructed a corpus by combining a Schwab S\&P 500 Index Fund prospectus (five densely written pages) with a deliberately injected irrelevant document describing various bird species. Five Schwab-specific questions were used against this combined corpus, which was chunked into 55 segments and clustered into three topical groups. Our methodology achieved an overall basic coverage of 66.5\%. Crucially, the cluster pertaining to 'bird species' exhibited a notably low coverage score of 43.2\%, falling significantly below the relevant Schwab-themed clusters (86.5\% and 87.4\%). This experiment validates that our coverage metrics are robust indicators for identifying semantically misaligned documents, enabling RAG system administrators to flag and remove off-topic content and maintain a more coherent and purposeful retrieval set.

\section{Discussion}

The validation experiments in real-world scenarios demonstrate the practical utility of our proposed RAG test coverage framework. By applying the methodology to diverse document sets, we confirmed its capability to provide actionable insights for improving test quality. The framework excels at identifying semantic blind spots in test questions and detecting irrelevant documents. By clustering document content and measuring the semantic proximity of test questions, we provide a quantifiable score that allows developers to systematically identify where testing efforts are lacking. The LLM-driven gap analysis further provides thematic suggestions for new questions, fostering an iterative improvement cycle.

Looking to the future, a critical area is optimizing test set efficiency. As test pools scale, manual review becomes expensive. An integrated efficiency metric would play a crucial role in flagging redundant questions, ensuring resources are focused on truly impactful questions. This would streamline test suite development, reduce human effort, and mitigate computational costs, aligning with our goal of continuously optimizing RAG test set quality.

\section{Limitations}

While our methodology offers significant advancements, it is important to acknowledge limitations. Our core approach relies on vector embeddings, which can struggle to perfectly capture the nuance, ambiguity, and context-dependency of human language. Issues like polysemy or negation might not be perfectly resolved, leading to potential inaccuracies. Furthermore, the results are directly influenced by the quality and inherent biases of the chosen embedding model.

Our reliance on automation also presents challenges. While the LOF algorithm effectively identifies distant questions, it cannot discern the reason for a question's irrelevance. Similarly, LLM-generated questions for gap analysis may not capture the most challenging or nuanced test cases that a human expert would devise. Therefore, the "human in the loop" remains crucial for interpreting outputs, making final decisions, and ensuring the test set reflects the most critical aspects of the RAG system’s intended use.

\section{Conclusion}
This work introduces a novel methodology for evaluating the quality of test sets in Retrieval-Augmented Generation (RAG) systems by borrowing and adapting concepts like code coverage from traditional software testing. By embedding both document chunks and test questions into a shared semantic space, we quantify test coverage using a diverse set of complementary metrics—including basic coverage, weighted coverage, and multi-cluster coverage techniques.

Our experiments across multiple domains demonstrate that this approach offers practical, actionable insights. In our product use case, we showed how coverage metrics could identify blind spots and improve test quality by guiding the addition of semantically aligned questions. Conversely, our evaluation using the Schwab S\&P 500 prospectus revealed how the same framework can flag semantically misaligned documents for removal, ensuring a cleaner and more relevant RAG knowledge base.

Ultimately, our coverage framework enables a more rigorous, scalable, and interpretable evaluation of test sets, complementing existing output-based RAG evaluation tools like RAGAS and ARES. By focusing on the input coverage problem, we provide an essential piece of the RAG evaluation pipeline—one that supports more reliable deployments and minimizes risk in high-stakes applications. We anticipate that this methodology can serve as both a diagnostic and an iterative improvement tool for any organization deploying RAG systems, and that future work may extend it to other stages of the RAG life-cycle such as usage monitoring and hallucination detection.

\bibliography{references}

\end{document}